\documentclass[a4paper]{article}
\usepackage[T1]{fontenc}
\usepackage[utf8]{inputenc}
\usepackage{amsmath}
\usepackage{amsfonts}
\usepackage{amssymb}
\usepackage{graphicx}
\usepackage{tikz}
\usepackage{mathtools}
\usepackage[top=3cm, bottom=3cm, left=3cm, right=3cm]{geometry}
\usepackage{calrsfs}
\usepackage{bbm}
\usepackage{algorithm}
\usepackage[noend]{algpseudocode}
\usepackage{float}
\DeclareMathAlphabet{\pazocal}{OMS}{zplm}{m}{n}

\makeatletter
\newcommand*\bigcdot{\mathpalette\bigcdot@{.5}}
\newcommand*\bigcdot@[2]{\mathbin{\vcenter{\hbox{\scalebox{#2}{$\m@th#1\bullet$}}}}}
\makeatother

\DeclareMathOperator*{\argmin}{argmin}

\newtheorem{theorem}{\indent Theorem}[section]

\newtheorem{definition-theorem}[theorem]{\indent Definition-Theorem}

\def \Z{\mathbb{Z}}

\def \P{\mathbb{P}}
\newcommand{\E}{\mathbb{E}}

\title{\bf A Numerical Study of the Time of Extinction in a Class of Systems of Spiking Neurons}
\author{C. Romaro$^1$, F.A. Najman$^2$ and M. André$^2$ \\ \textit{Departamento de Física$^1$, and Instituto de Matemática e Estatística$^2$} \\ \textit{Universidade de São Paulo.}}
\date{March 2019}

\date{\today}

\begin{document}

\maketitle

\section*{Abstract}
In this paper we present a numerical study of a mathematical model of spiking neurons introduced by Ferrari et al. in \cite{ferrari2018phase}. 
In this model we have a countable number of neurons linked together in a network, each of them having a membrane potential taking value in the integers, and each of them spiking over time at a rate which depends on the membrane potential through some rate function $\phi$. Beside being affected by a spike each neuron can also be affected by leaking. At each of these leak times, which occurs for a given neuron at a fixed rate $\gamma$, the membrane potential of the neuron concerned is spontaneously reset to $0$. A wide variety of versions of this model can be considered by choosing different graph structures for the network and different activation functions. It was rigorously shown that when the graph structure of the network is the one-dimensional lattice with a hard threshold for the activation function, this model presents a phase transition with respect to $\gamma$, and that it also presents a metastable behavior. By the latter we mean that in the sub-critical regime the re-normalized time of extinction converges to an exponential random variable of mean 1. It has also been proven that in the super-critical regime the renormalized time of extinction converges in probability to 1. Here, we investigate numerically a richer class of graph structures and activation functions. Namely we investigate the case of the two dimensional and the three dimensional lattices, as well as the case of a linear function and a sigmoid function for the activation function. We present numerical evidence that the result of metastability in the sub-critical regime holds for these graphs and activation functions as well as the convergence in probability to $1$ in the super-critical regime.

\section{Introduction}
Informally the model we consider is as follows. $I$ is a countable set representing the neurons, and to each $i \in I$ is associated a set $\mathbb{V}_i$ of \textit{presynaptic neurons}. If you consider the elements of $I$ as vertices, and draw and edge from $j$ to $i$ whenever $j \in \mathbb{V}_i$, then you obtain the graph structure of the network. The \textit{membrane potential} of neuron $i$ is a stochastic process denoted $(X_i(t))_{t \geq 0}$ taking value in the set $\mathbb{N}$ of non-negative integers. Moreover, we associate to each neuron a Poisson process $(N^{\dagger}_i(t))_{t \geq 0}$ of some parameter $\gamma$, representing the \textit{leak times}. At any of these leak times the membrane potential of the neuron concerned is reset to $0$. Another point process $(N_i(t))_{t \geq 0}$ representing the \textit{spiking times} is also associated to each neuron, which rate at time $t$ is given by $\phi (X_i(t))$, where $\phi$ is the rate function. Whenever a neuron spikes its membrane potential is also reset to $0$ and the membrane potential of all of its post-synaptic neurons is increased by one (i.e. the neurons of the set $\{ j : i \in  \mathbb{V}_j\}$). All the point processes involved are assumed to be are mutually independent.

\vspace{0.4 cm}

More formally, beside asking that $(N^{\dagger}_i(t))_{t \geq 0}$ be a Poisson process of some parameter $\gamma$, this is the same as saying that $(N_i(t))_{t \geq 0}$ is the point process characterized by the following equation $$\E (N_i (t)- N_i (s)|\mathcal{F}_s) = \int_s^t \E (\phi (X_i(u))|\mathcal{F}_s)du $$ where $$X_i(t) = \sum_{j \in \mathbb{V}_i}\int_{]L_i(t),t[}dN_j(s),$$ $L_i(t)$ being the time of the last event affecting neuron $i$ before time $t$, that is, $$L_i(t) = \sup \Big\{s \leq t : N_i(\{s\}) = 1 \text{ or } N^\dagger_i(\{s\}) = 1\Big\}.$$ $(\mathcal{F}_t)_{t\geq 0}$ is the standard filtration for the point processes involved here. See \cite{ferrari2018phase} for more details.

\vspace{0.4 cm}

In \cite{ferrari2018phase}, \cite{andre2019meta}, and \cite{andre2019determ} a specific version of the model above was studied. The graph structure chosen there was the one-dimensional lattice, i.e. $I = \Z$ with $\mathbb{V}_i = \{i-1,i+1\}$, moreover the activation function was chosen to be a hard threshold of the form $\phi(x) = \mathbbm{1}_{x > 0}$. In such a paradigm the rate of the point processes representing spiking times can only take two values: $0$ or $1$, depending on whether the membrane potential is positive or null. In this context whenever the membrane potential of a neuron is different from $0$ we say that the neuron is \textit{active}, otherwise we say that it is \textit{quiescent}. More generally we will only consider here function $\phi$ satisfying $\phi(0) = 0$ and $\phi(x) > 0$ for $x > 0$, so that we will keep this distinction between active and quiescent neurons.

\vspace{0.4 cm}

In \cite{ferrari2018phase} it was proven that in the case of the one-dimensional lattice with hard threshold the following theorem holds

\begin{theorem} \label{thm:phasetransition}
Suppose that for any $i \in \Z$ we have $X_i(0) \geq 1$. There exists a critical value $\gamma_c$ for the parameter $\gamma$, with $0 < \gamma_c < \infty$, such that for any $i \in \Z$

$$\P \Big( N_i([0,\infty[) \text{ } < \infty \Big) = 1 \text{ if } \gamma > \gamma_c$$

and

$$ \P \Big( N_i([0,\infty[) \text{ } = \infty \Big) > 0 \text{ if } \gamma < \gamma_c.$$

\end{theorem}

In words there is a critical point for the parameter $\gamma$, such that below this critical point each neuron stays active forever (with positive probability), and above it each neuron becomes quiescent if you wait long enough.

\vspace{0.4 cm}

The process as a whole of course never dies completely because of the fact that there is an infinite number of neurons, so that it doesn't makes sense to consider the extinction time. Nonetheless nothing prevents us to consider a finite version of this model. Suppose we're still in the case  $\mathbb{V}_i = \{i-1,i+1\}$ and $\phi(x) = \mathbbm{1}_{x > 0}$ and for any $N \geq 0$ consider the system defined on the finite set $I_N = \{-N,\ldots N\}$ instead of the whole lattice. Then you can define the process $(\xi_N(t))_{t \geq 0}$ of the spiking rates of the system, that is, the process taking value in $\{0,1\}^{I_N}$ defined by ${\xi_N(t)}_i = \mathbbm{1}_{X_i(t) > 0}$. This is a Markov process, belonging to the category of interacting particle systems, and by classical results on Markov processes we know that it needs to reach the state $0^{I_N}$ - where all neurons are quiescent - in finite time. We can therefore consider the extinction time of this finite model, which we denote $\sigma_N$, and it is natural to ask about its distribution in each of the two phases distinguished by the theorem above. It was proven in \cite{andre2019meta} that the following holds.

\begin{theorem} \label{thm:metamodel}
There exists $\gamma'_c$ such that if $\gamma < \gamma'_c$, then we have the following convergence

$$\frac{\sigma_N}{\E (\sigma_N)} \overset{\mathcal{L}}{\underset{N \rightarrow \infty}{\longrightarrow}} \mathcal{E} (1).$$

In words, the re-normalized time of extinction converges in distribution to an exponential random variable of mean 1.
\end{theorem}

We know that $\gamma'_c < \gamma_c$, and the fact that the theorem is stated for some $\gamma'_c$ and not for the critical value $\gamma_c$ comes from essentially technical reasons related to the way the proof is built. We believe that this metastable result holds for the whole sub-critical region but it is not yet proven. Moreover, it was proven in \cite{andre2019determ} that we also have the following.

\begin{theorem} \label{thm:deterministic}
Suppose that $\gamma > 1$. Then the following convergence holds

$$\frac{\sigma_N}{\E (\sigma_N)}  \overset{\P}{\underset{N \rightarrow \infty}{\longrightarrow}} 1.$$
\end{theorem}

We know that $\gamma_c < 1$, so that the result concerns a sub-region of the super-critical region, and as for the previous result, while to the best of our knowledge the result has been proven only for $\gamma > 1$, it is reasonable to expect that it holds in the whole super-critical region.

\vspace{0.4 cm}
The choice of a one-dimensional lattice for the graph of interaction and of a hard threshold for the activation function were initially essentially motivated by mathematical conveniency, and we're interested in checking that the results hold for a richer class of instantiations of the model. In this paper we investigate numerically cases for which we don't have yet any rigorous result. Namely, we investigate higher dimension lattices $\Z^d$ ($d = 2$ and $d = 3$) to show that the results related to the asymptotical distribution of the extinction time stated for the one-dimensional case in Theorem \ref{thm:metamodel} and Theorem \ref{thm:deterministic} remain valid for these graphs. The choice of such structure for the graph of interaction is common in the literature that is concerned with mathematical modeling of neural networks (see for example \cite{miranda1991self} and \cite{makarenkov1991self}), and it is justified by the fact that highly connected cortical regions, such as specific regions of the primate visual cortex \cite{rockland1983intrinsic}\cite{yoshioka1992intrinsic}, present some similarity with multidimensional lattices \cite{sporns2004motifs}. We also investigate the effect of changing the activation function to a linear function and to a sigmoid function. Both activation functions have been used in some form in stochastic models of biological neural networks (see for example \cite{brochini2016phase} and \cite{kinouchi2019stochastic}) as well as in artificial neural networks (see for example \cite{maass1997networks}).

\vspace{0.4 cm}

This paper is organized as follows. In Section \ref{algo} we give a description of the algorithm used for the simulations. In Section \ref{models} we describe the specific instantiations of the model that we investigate. In Section \ref{results} we present the results of our simulations. Finally in Section \ref{discussion} we discuss these results.

\section{Simulation algorithm}
\label{algo}

All simulations were done in Python. The algorithm used to simulate our system of spiking neurons can be informally described as follows:

\begin{itemize}
\item{\textbf{Initial configuration:}}
The network start with all neurons actives (by default membrane potential equal $1$), and for each neuron an initial spiking value is sampled from an exponential random variable of parameter $1$ (corresponding to the first atom of $(N_i (t))_{t \geq 0}$) and an initial leaking time value is sampled for each neuron from an another independent exponential random variable of parameter $\gamma$ (corresponding to the first atom of $(N_i^\dagger (t))_{t \geq 0}$).

\item{\textbf{Interaction:}}
The current time is set to be the smallest of the previously sampled values, the neuron $i$ associated with this value is found and the value of its membrane potential is set to $0$. In case the event considered is a spike the membrane potential of the neurons in the set $\{j: i \in \mathbb{V}_j\}$ (the set of post-synaptic neurons) is increased by one. The membrane potential of neuron $i$ being equal to $0$ we set the next spiking time for neuron $i$ to be infinite until further notice. If the event was a leaking then we sample an exponential random variable of parameter $\gamma$ and add it to the current time to get the next leaking time for neuron $i$. If the event was a spike then we sample an exponential random variable of parameter $\phi(X_j)$ for all post-synaptic neuron $j$ and add this value to the current time to get the next spiking time for these neurons.

\item{\textbf{Stopping condition:}}
The previous operation is iterated until all neurons are quiescent.
\end{itemize}

\vspace{0.4 cm}

More formally, our simulation algorithm can be described by the following pseudo-algorithm.

\vspace{0.4 cm}

\begin{algorithm}[H]
\caption{Simulate the system of spiking neurons and return the extinction time}\label{alg:sim}
\begin{algorithmic}[1]

\State $I$ the (finite) set of neurons.
\State $\phi$ the activation function.
\State $\gamma$ the rate of the leaking point processes.
\State $t$ the current time.
\State $\mathbb{V}_i$ the set of presynaptic neurons for neuron $i$.
\State $X_i$ the membrane potential of neuron $i$ at the current time.
\State $\sigma^\dagger_i$ the time of the next leaking for neuron $i$.
\State $\sigma_i$ the time of the next spike for neuron $i$.

\vspace{0.6 cm}

INITIALIZATION

\vspace{0.4 cm}

\State $t \gets 0$

\For{each $i$ in $I$}
    \State $X_i \gets 1$
\EndFor

\For{each $i$ in $I$}
    \State $\sigma^\dagger_i \gets \mathcal{E} (\gamma)$ \Comment{$\mathcal{E}$ denotes the realization of an exponential random variable}
    \State $\sigma_i \gets \mathcal{E} (\phi(X_i))$
\EndFor

\vspace{0.6 cm}

SIMULATION

\vspace{0.4 cm}

\While{$\sum_{i \in I} X_i \neq 0$}

    \vspace{0.2 cm}    

    \State $MinLeaking \gets \min_{i \in I} \sigma^\dagger_i$
    \State $MinSpiking \gets \min_{i \in I} \sigma_i$
    
    \vspace{0.2 cm}    

    \If{$MinLeaking < MinSpiking$}
        \State $t=MinLeaking$
        \State $i \gets \argmin_{j \in I} \sigma^\dagger_j$
        \State $X_i \gets 0$
        \State $\sigma_i \gets \infty$
        \State $\sigma^\dagger_i \gets t + \mathcal{E} (\gamma)$
    \Else
        \State $t=MinSpiking$
        \State $i \gets \argmin_{j \in I} \sigma_j$
        \State $X_i \gets 0$ 
        \State $\sigma_i \gets \infty$
        \For{each $j$ such that $i \in \mathbb{V}_j$}
            \State $X_j \gets X_j + 1$
            \State $\sigma_j \gets t + \mathcal{E} (\phi(X_j))$
        \EndFor
       
\EndIf
\EndWhile

\vspace{0.4 cm}

\State $\sigma_N \gets t$ 
\State \Return $\sigma_N$
\end{algorithmic}
\end{algorithm}

\newpage

\section{Models investigated}
\label{models}

In this section we specify the structure of the graph of interaction and the activation function we're interested in.

\subsection{Multi-dimensional lattices}

For the graph of the network we consider the lattices $\Z^1$, $\Z^2$ and $\Z^3$. For any $d \in \{1,2,3\}$, let $\| \cdot \|$ be the norm on $\Z^d$ given for any $j\in \Z^d$ by $$\|j\| = \sum_{k=1}^d |j_k|,$$ where $j_k$ is the k-th coordinate of $j$. The structure of the network is then given by $I = \Z^d$ and $\mathbb{V}_i = \{j \in I^d : \|i-j\| = 1\}$ for $i \in I$.

\vspace{0.4 cm}

Notice that by defining the set of presynaptic neurons as the set of the nearest neighbours we actually have $j \in \mathbb{V}_i$ if and only if $i \in  \mathbb{V}_j$. In other words, for a given neuron the set of the presynaptic neurons and the set of the postsynaptic neurons are equal. For this specific choice the graph of interaction is therefore actually undirected.

\begin{figure}[H]
        \center{\includegraphics[width=9cm]
        {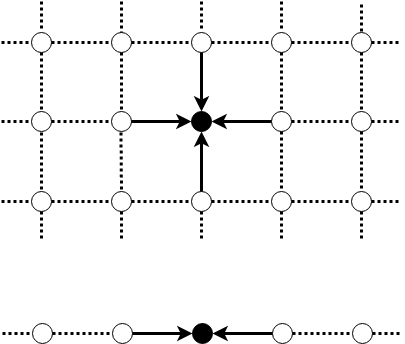}}
        \caption{\label{lattices} One-dimensional and two-dimensional lattices. A directed arrow is drawn toward the black neuron from each of its presynaptic neurons.}
\end{figure}

\vspace{0.4 cm}

\subsection{Linear and sigmoid activation functions}

The activation function considered in \cite{andre2019meta}, \cite{ferrari2018phase} and \cite{andre2019determ} was the hard threshold $\phi(x) = \mathbbm{1}_{x > 0}$. This choice is convenient mathematically as the system then becomes an additive interacting particle system where any neuron can only have two possible values for the spiking rate at any time: $0$ or $1$. Nonetheless, from a biological point of view, a hard threshold is a rough choice, and we would like to consider smoother options.

\vspace{0.4 cm}

The first option we consider is a linear function of the simplest form: $\phi(x) = x$. 

\vspace{0.4 cm}

The second option we consider is a somewhat more sophisticated sigmoid function of the following form

$$\phi(x) =
\begin{cases}
    (1 + e^{-3x + 6})^{-1}& \text{ if } x > 0 ,\\
    0              & \text{ if } x = 0.
\end{cases}$$

\vspace{0.4 cm}

Notice that we need to have $\phi (0) = 0$, in order to avoid spontaneous spiking (neuron with null membrane potential that spikes nonetheless). This is the reason why we force this value for the sigmoid function.

\section{Results}
\label{results}

We run simulations for instantiations of the system of spiking neurons consisting of all the possible combinations between the graphs and activation functions described above.

\subsection{Simulations with a fixed number of neurons }

\subsubsection{Multidimensional lattices and hard threshold}

For each of the three lattices the Algorithm \ref{alg:sim} described in Section \ref{algo} was run, with an activation function of the form $\phi(x) = \mathbbm{1}_{x > 0}$. Each of the simulations were run for two different values of $\gamma$, 10,000 times for each of these values, using a number of neurons of the order of 100. The mean of the time of extinction $\sigma_N$ was then computed using these data, and used to build the re-normalized histogram in each of these cases. The exact values for the size of the network and for the parameter $\gamma$ can be found in Table \ref{parametershard}.

\vspace{0.4 cm}

\begin{table}[h]
\center{\begin{tabular}{|c|c|c|c|}
\hline
Lattice                  & Number of neurons & Value of $\gamma$ & \multicolumn{1}{l|}{Figure} \\ \hline
$\Z$                     & 101               & 0.34                  & \ref{hist_hardthre_sub}     \\
                         &                   & 0.85                  & \ref{hist_hardthre_sup}     \\ \hline
$\Z^2$                   & 121               & 1.25                  & \ref{hist_hardthre_sub}     \\
                         &                   & 5.00                  & \ref{hist_hardthre_sup}     \\ \hline
$\Z^3$                   & 125               & 1.80                  & \ref{hist_hardthre_sub}     \\
                         &                   & 6.00                  & \ref{hist_hardthre_sup}     \\ \hline
\end{tabular}
\caption{Values of the total number of neurons and of the parameter $\gamma$ used in the simulation for each of the three lattices.} \label{parametershard}}
\end{table}

\vspace{0.4 cm}

\begin{figure}[H]
        \center{\includegraphics[width=12 cm]
        {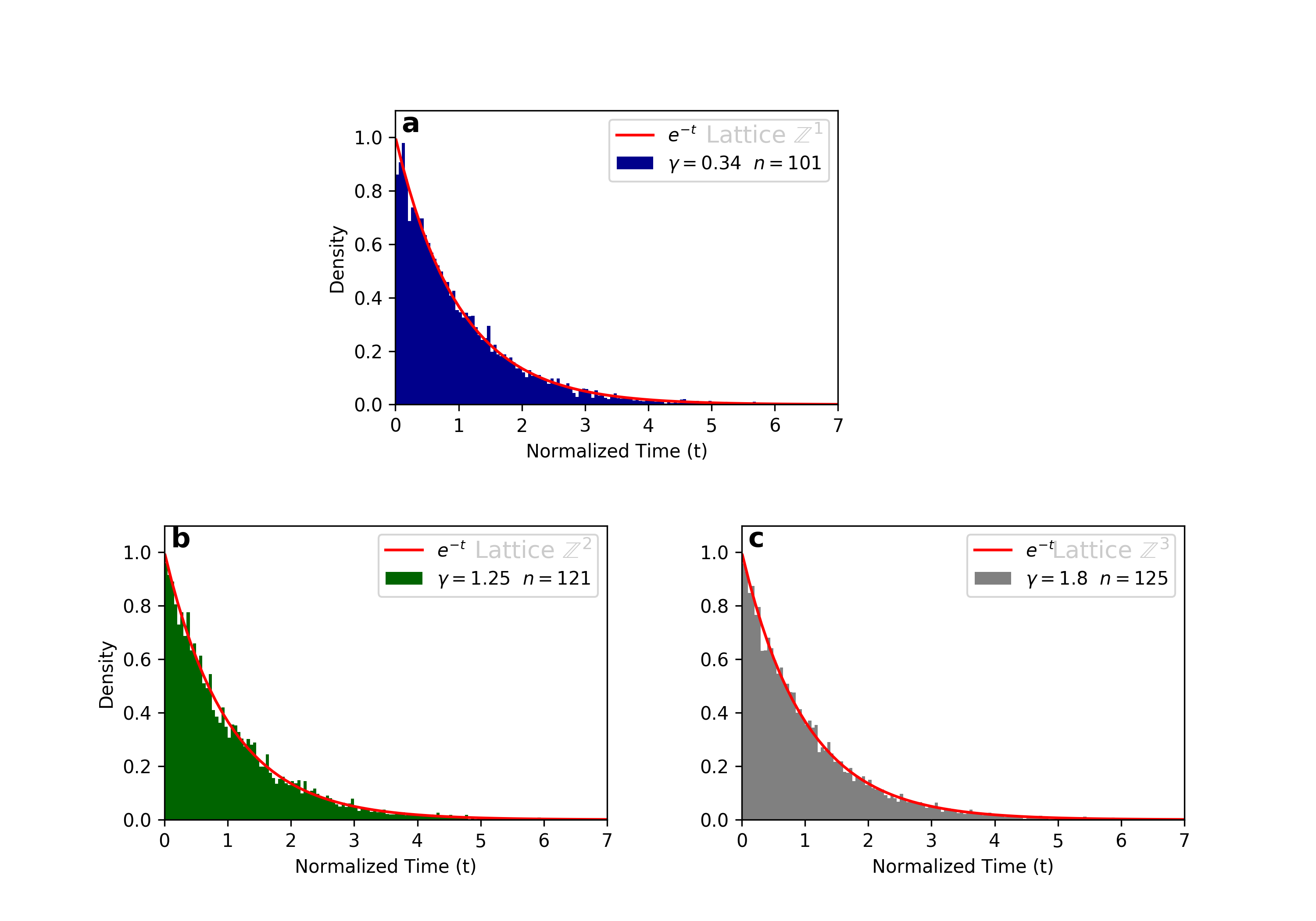}}
        \caption{\label{hist_hardthre_sub} Histogram of the re-normalized time of extinction $\sigma_N$ for small values of gamma, and an activation function of the form $\phi(x) = \mathbbm{1}_{x > 0}$. In \textbf{a}, \textbf{b} and \textbf{c} the blue, green and gray bars are the histograms for the time of extinction in the one-dimensional lattice, two-dimensional lattice and three-dimensional lattice respectively. The red line is the exponential function $t \mapsto e^{-t}$, which corresponds to the density of an exponential law of parameter 1. The parameter $n$ corresponds to the number of neurons.}
\end{figure}

\begin{figure}[H]
        \center{\includegraphics[width=12cm]
        {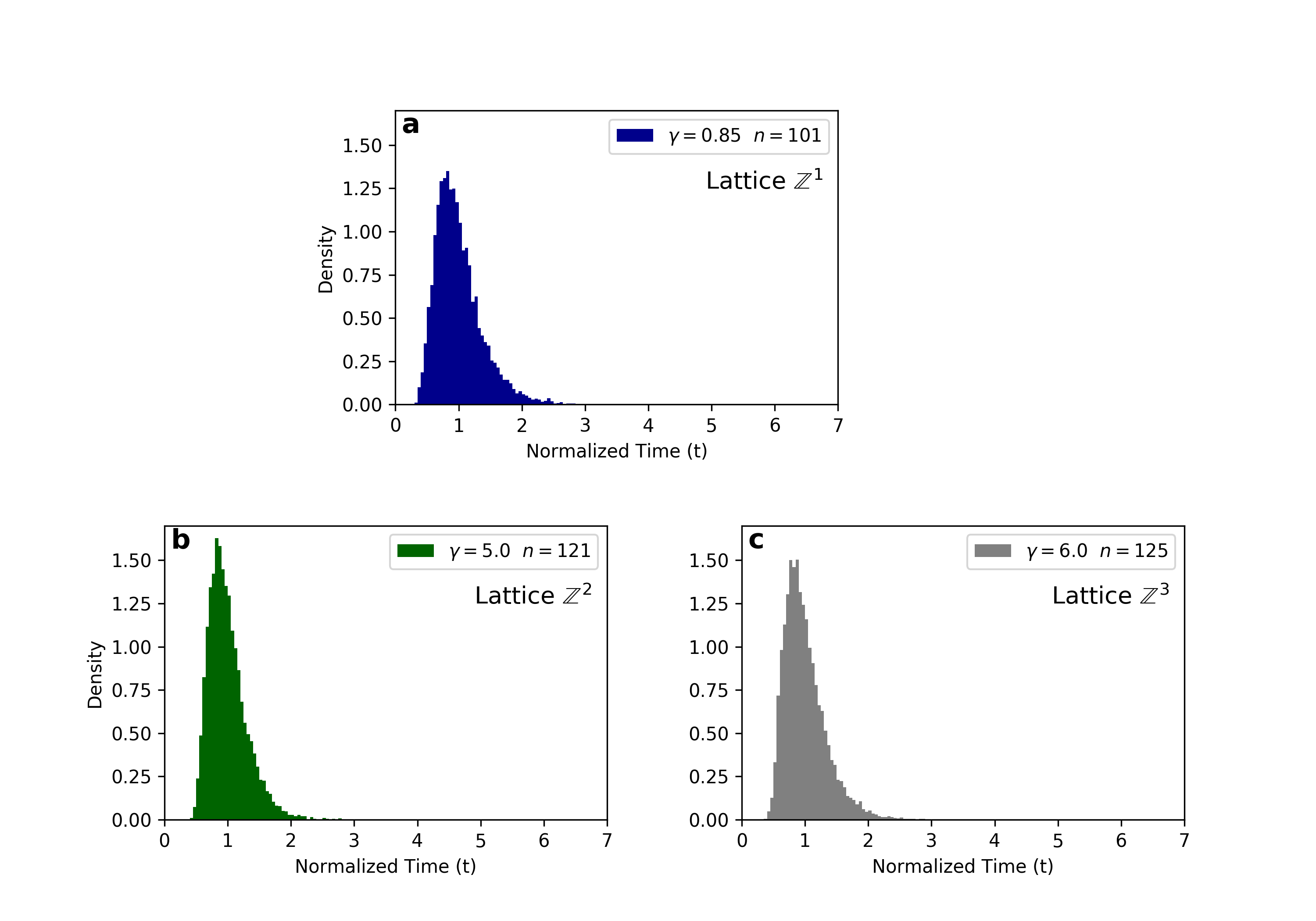}}
        \caption{\label{hist_hardthre_sup} Histogram of the re-normalized time of extinction $\sigma_N$ for high values of gamma, and an activation function of the form $\phi(x) = \mathbbm{1}_{x > 0}$. In \textbf{a}, \textbf{b} and \textbf{c} the blue, green and gray bars are the histograms for the time of extinction in the one-dimensional lattice, two-dimensional lattice and three-dimensional lattice respectively. The parameter $n$ corresponds to the number of neurons.}
\end{figure}

The resulting histograms are presented in Figure \ref{hist_hardthre_sub} and \ref{hist_hardthre_sup}.

\vspace{0.4 cm}

\subsubsection{Multidimensional lattices, linear function and sigmoid function}

The routine described above was repeated with the two other activation functions. Only the values of $\gamma$ change, which was necessary as changing the activation function must change the critical value of the system. These values are given in Table \ref{parameterslinsig}.

\begin{table}[!h]
\center{\begin{tabular}{|c|c|c|c|c|}
\hline
Lattice                  & Number of neurons & Activation function & Value of $\gamma$ & \multicolumn{1}{l|}{Figure}  \\ \hline
$\Z$                     & 101               & Linear              & 0.42               & \ref{histlin}                                         \\
                         &                   & Sigmoid             & 0.028              & \ref{histsig}                \\
                         &                   & Linear              & 1                  & \ref{histlin}                \\
                         &                   & Sigmoid             & 0.85               & \ref{histsig}                \\ \hline
$\Z^2$                   & 121               & Linear              & 1.70               & \ref{histlin}                                         \\
                         &                   & Sigmoid             & 0.2                & \ref{histsig}                \\ 
                         &                   & Linear              & 5.00               & \ref{histlin}                \\
                         &                   & Sigmoid             & 1.7                & \ref{histsig}                \\ \hline
$\Z^3$                   & 125               & Linear              & 1.90               & \ref{histlin}                                         \\
                         &                   & Sigmoid             & 0.09               & \ref{histsig}                \\ 
                         &                   & Linear              & 6.00               & \ref{histlin}                \\ 
                         &                   & Sigmoid             & 1.8                & \ref{histsig}                \\ \hline
\end{tabular}
\caption{Values of the total number of neurons and of the parameter $\gamma$ used in the simulation for each of the three lattices.} \label{parameterslinsig}}
\end{table}

\begin{figure}[H]
        \center{\includegraphics[width=15 cm]
        {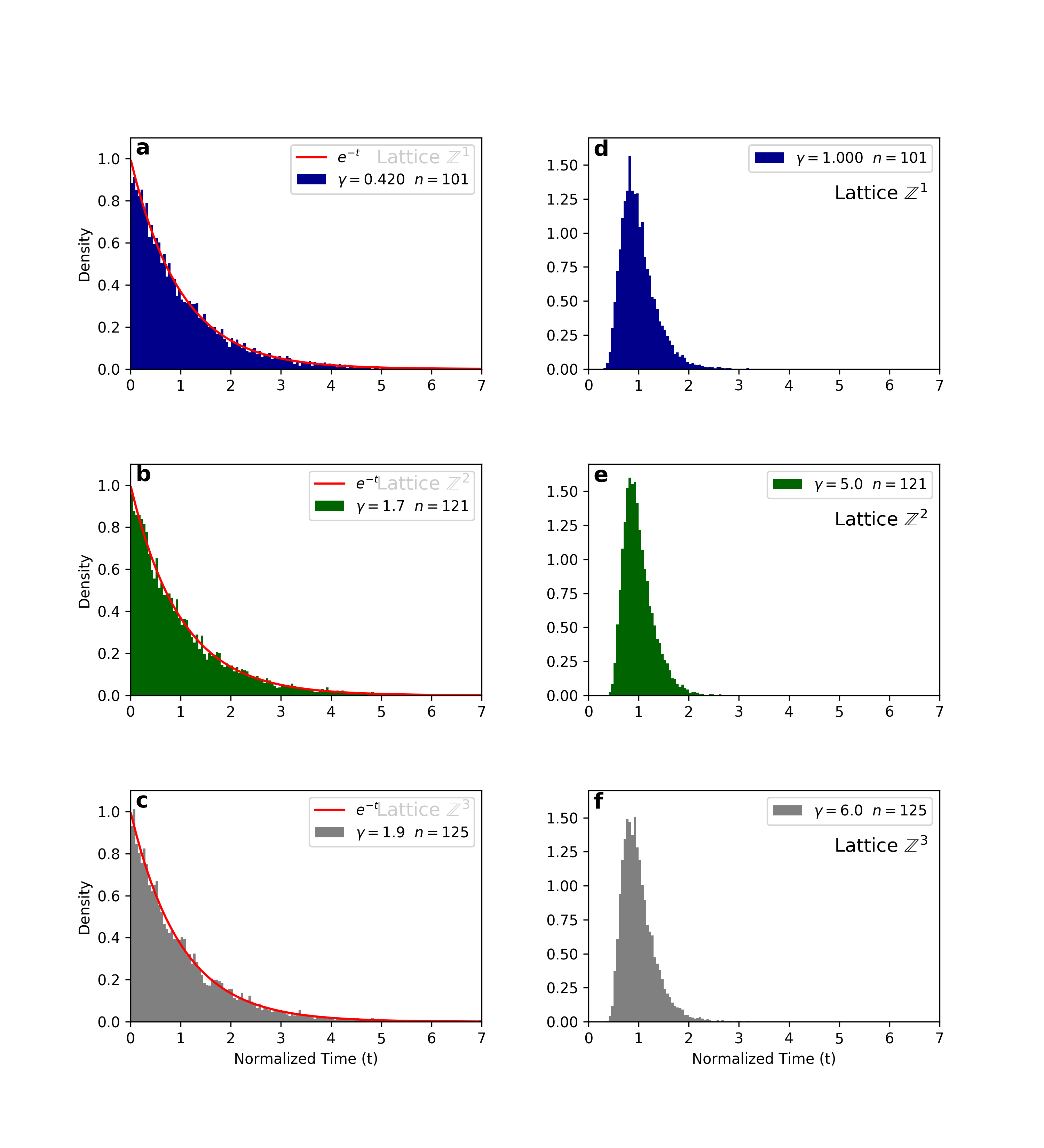}}
        \caption{\label{histlin} Histogram of the re-normalized time of extinction $\sigma_N$ for a linear activation function for each of the three lattices.  On the left side are the histograms for small values of $\gamma$ and on the right side the histograms for high values of $\gamma$. The red line on the left side is the exponential function $t \mapsto e^{-t}$, which corresponds to the density of an exponential law of parameter 1. The parameter $n$ corresponds to the number of neurons.}
\end{figure}

\begin{figure}[H]
        \center{\includegraphics[width=15 cm]
        {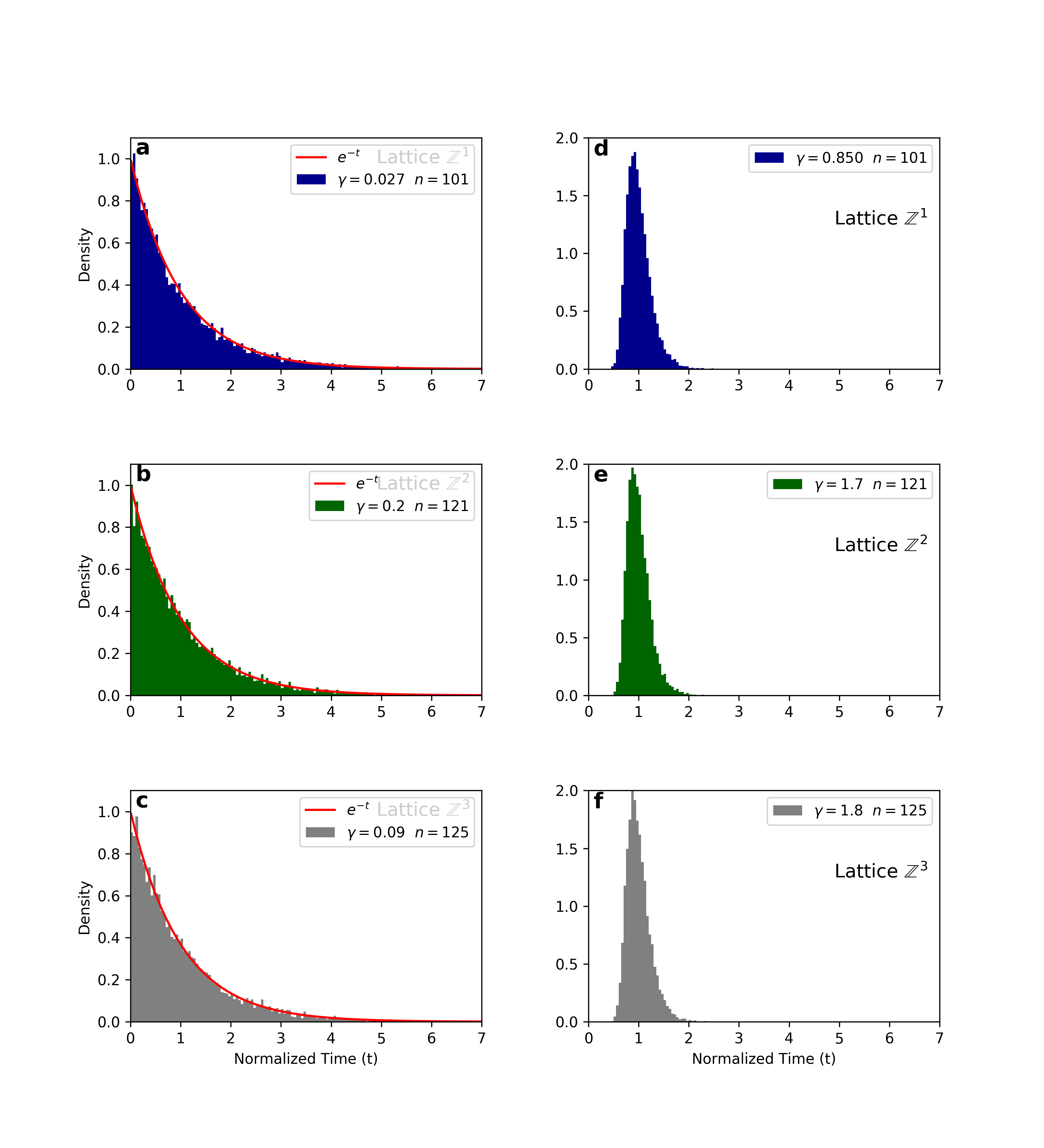}}
        \caption{\label{histsig} Histogram of the re-normalized time of extinction $\sigma_N$ for a sigmoid activation function for each of the three lattices. On the left side are the histograms for small values of $\gamma$ and on the right side the histograms for high values of $\gamma$. The red line on the left side is the exponential function $t \mapsto e^{-t}$, which corresponds to the density of an exponential law of parameter 1. The parameter $n$ corresponds to the number of neurons.}
\end{figure}

The resulting histograms are presented in figures \ref{histlin} and \ref{histsig}.

\subsection{Simulations for a varying number of neurons}

To further investigate the behavior of the time of extinction in the super-critical regime, we've run a set of simulation for a (fixed) high value of $\gamma$ and for a varying number of neurons. Each element of the set consists in 1000 repetitions with $\gamma = 4$. The value of the size of the network varies from $11$ to $2000$. For each of these values we've estimated the mean and variance of the extinction time $\sigma_N$, and the variance of the re-normalized extinction time $\sigma_N / \E(\sigma_N)$. These simulations have been done in the one-dimensional lattices for all of the three activation functions

\vspace{0.4 cm}

\begin{figure}[H]
        \center{\includegraphics[width=12cm]
        {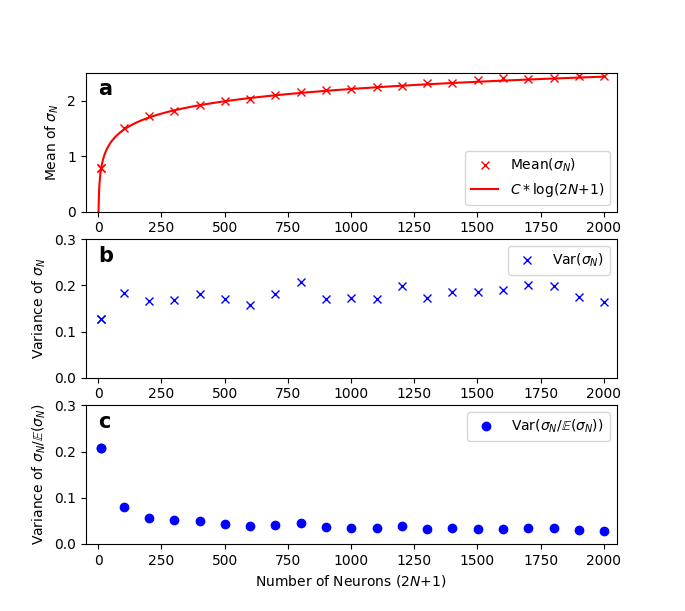}}
        \caption{\label{varying_hard} Mean and variance of $\sigma_N$ and variance of the renormalized extinction time $\sigma_N / \E (\sigma_N)$ for a linear activation function. In \textbf{a} the red dots represents the values of the estimated mean of $\sigma_N$ for varying numbers of neurons and the red line a logarithmic function fitted over the values of the mean (C = 0.32). In \textbf{b} the blue crosses represent the variance of $\sigma_N$ as the number of neurons increases. The blue dots in \textbf{c} represent the variance of the renormalized extinction time $\sigma_N / \E (\sigma_N)$ for a varying number of neurons. All simulations were run with $\gamma = 4$.}
\end{figure}

\begin{figure}[H]
        \center{\includegraphics[width=10cm]
        {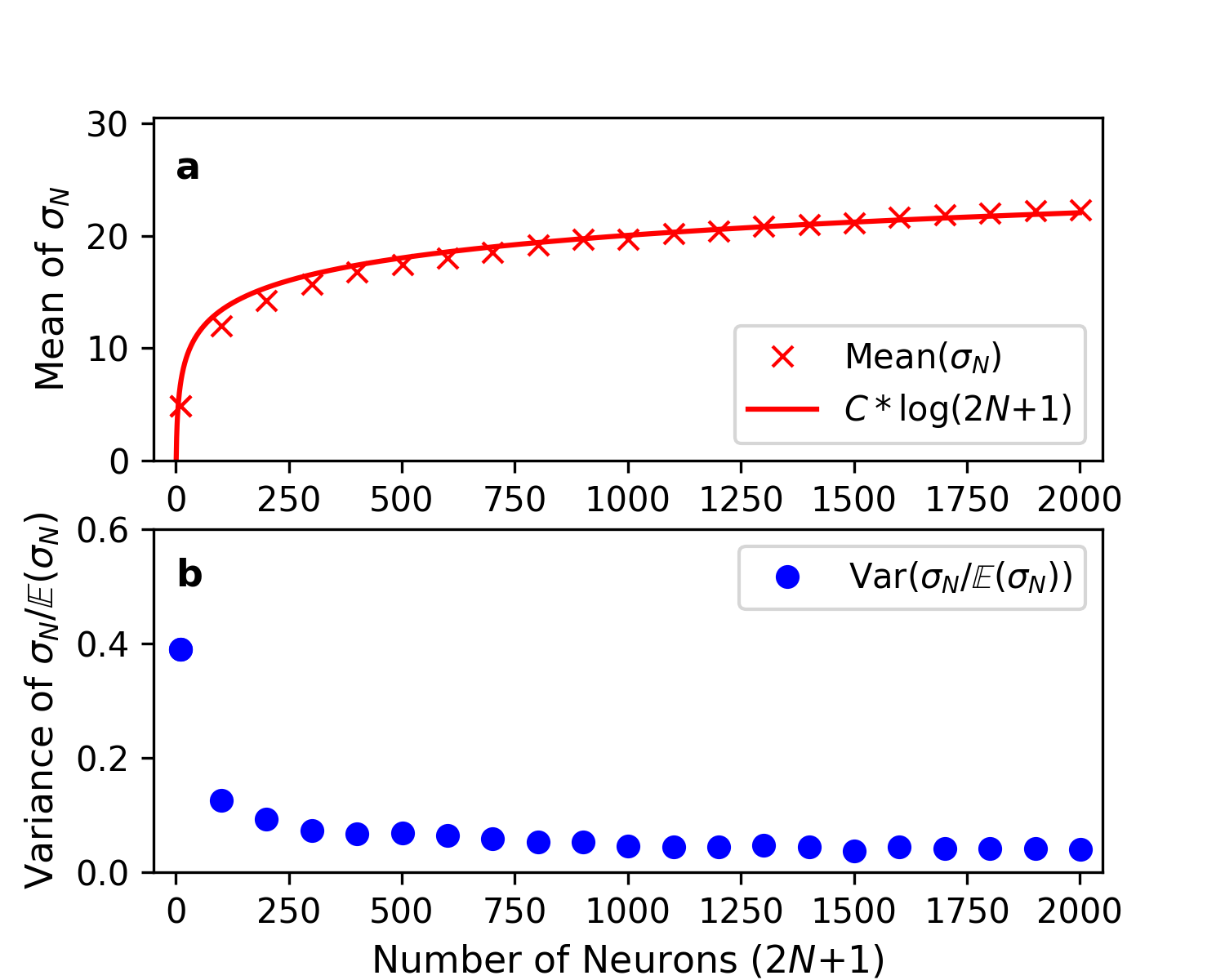}}
        \caption{\label{varying_lin} Mean of $\sigma_N$ and variance of the renormalized extinction time $\sigma_N / \E (\sigma_N)$ for a linear activation function. In \textbf{a} the red dots represents the values of the estimated mean of $\sigma_N$ for varying numbers of neurons and the red line a logarithmic function fitted over the values of the mean. The blue dots in \textbf{b} represent the variance of the renormalized extinction time $\sigma_N / \E (\sigma_N)$ for a varying number of neurons. All simulations were run with $\gamma = 4$.}
\end{figure}

\begin{figure}[H]
        \center{\includegraphics[width=10cm]
        {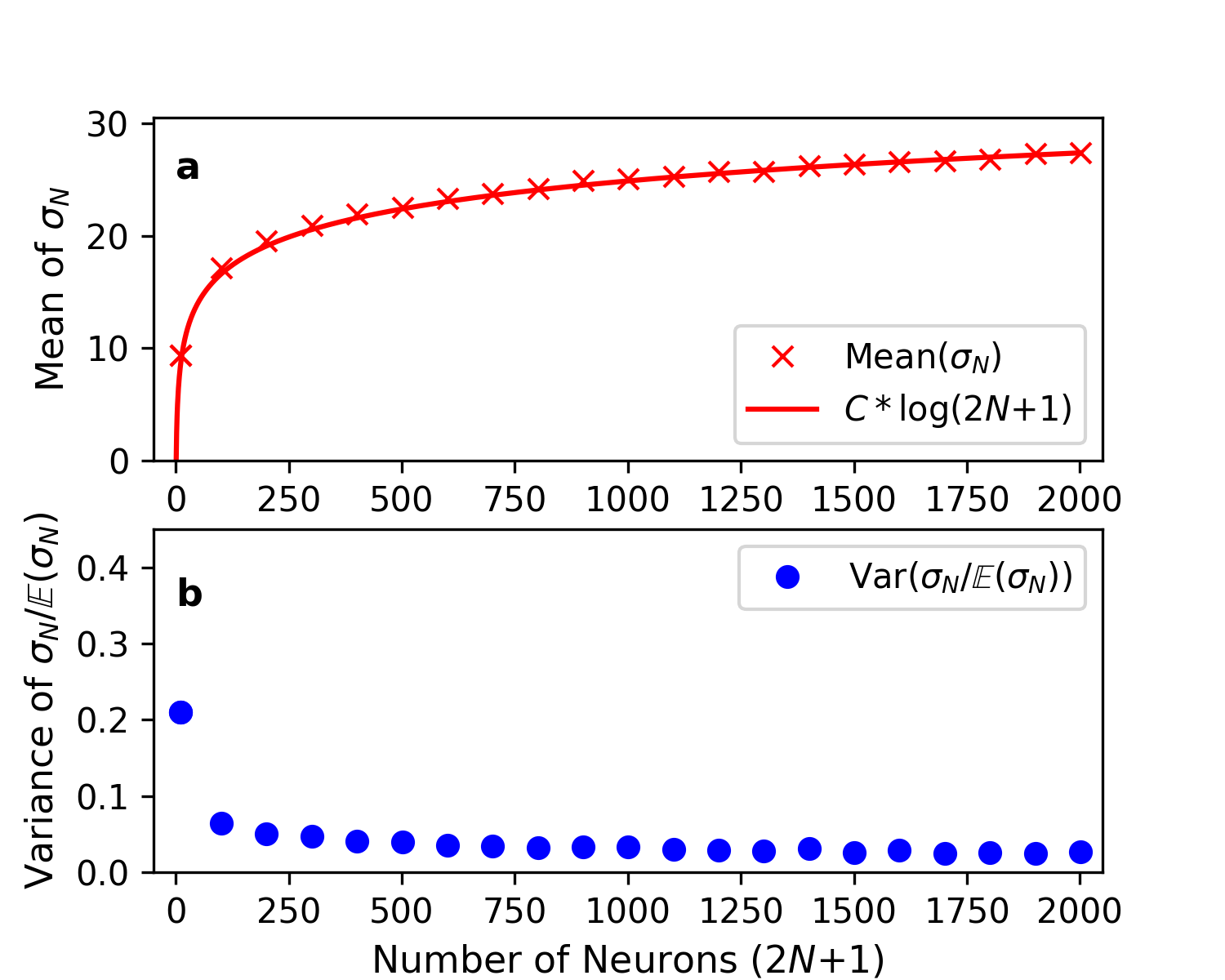}}
        \caption{\label{varying_sig} Mean of $\sigma_N$ and variance of the renormalized extinction time $\sigma_N / \E (\sigma_N)$ for a sigmoid activation function. In \textbf{a} the red dots represents the values of the estimated mean of $\sigma_N$ for varying numbers of neurons and the red line a logarithmic function fitted over the values of the mean. The blue dots in \textbf{b} represent the variance of the renormalized extinction time $\sigma_N / \E (\sigma_N)$ for a varying number of neurons. All simulations were run with $\gamma = 4$.}
\end{figure}

The results of these simulations are presented in Figure \ref{varying_hard}, Figure \ref{varying_lin} and Figure \ref{varying_sig}.

\vspace{0.4 cm}

\section{Discussion}
\label{discussion}

\subsection{Sub-critical regime}

The histogram built from the simulations for which $\gamma$ is small (Figure \ref{hist_hardthre_sub} and left side of Figure \ref{histlin} and Figure \ref{histsig}) closely approximates the density of a mean $1$ exponential random variable. The fact that the result of the simulations remains identical in all the cases investigated (dimension one, two and three, with hard-threshold, linear function and sigmoid function) suggests that Theorem \ref{thm:metamodel} doesn't merely hold for the specific instantiation of the model for which is was proven, but for a wide class of systems.

\vspace{0.4 cm}

In the one dimensional case with hard threshold the fact that the histogram approximates the density of an exponentially distributed random variable is of course not a surprise as this what Theorem \ref{thm:metamodel} predicts asymptotically. nonetheless it gives us evidences that the approximation by an exponential law holds for relatively small networks (in the simulation concerned the number of neuron in the system is $101$). The number of neurons in animals varies from hundreds \cite{white1986structure} to billions \cite{lent2012many}, so that in our model the approximation by an exponential law is observed for all possible biologically realistic number of neurons. This indicates that, in networks where the randomness of the connections between neurons is not of interest, the model presented here might be an interesting choice for the investigation of metastable behaviors in dense cortical regions.

\vspace{0.4 cm}

\subsection{Super-critical regime}
The histograms built from the simulations with high values of $\gamma$ are visibly not approximating any exponential. Instead they show a distribution that is reminiscent of a gamma distribution with mass concentrated around 1 (See Figure \ref{hist_hardthre_sup} and right side of Figure \ref{histlin} and Figure \ref{histsig}).

\vspace{0.4 cm}

Moreover the evolution of the variance of the renormalized time of extinction (last graph in Figure \ref{varying_hard}, Figure \ref{varying_lin} and Figure \ref{varying_sig}) is seemingly converging toward $0$ as the number of neurons grows for all of the three instantiations investigated. 

\vspace{0.4 cm}

For the simulation of the system with hard-threshold, these facts are not a surprise neither, as this is what Theorem \ref{thm:deterministic} predicts. Again the fact that the simulations of the systems with a linear and a sigmoid function for $\phi$ show similar results suggests that Theorem \ref{thm:deterministic} isn't only satisfied for $\phi(x) = \mathbbm{1}_{x > 0}$, but for a wide class of activation functions.

\vspace{0.4 cm}

Moreover the first graph in Figure \ref{varying_hard}, Figure \ref{varying_lin} and Figure \ref{varying_sig} gives us strong evidence that in each of the three instantiations the expectation of the time of extinction grows approximately like a logarithm (up to a multiplicative constant) with respect to the number of neurons. This fact is interesting in itself as the proof of Theorem \ref{thm:deterministic} (which can be found in \cite{andre2019determ}) relies on the fact that for the hard-threshold instantiation of the model we have  the following convergence in the super-critical region (at least for $\gamma > 1$)

\begin{equation}
\label{expectlog}
\frac{\E (\tau_N)}{\log (2N + 1)} \underset{N \rightarrow \infty}{\longrightarrow} C,
\end{equation}

where $C$ is a strictly positive (and finite) constant. This is an additional hint that the behavior of the time of extinction should be qualitatively identical in the super-critical region as well for any of the choices we proposed here for the activation function.

\newpage

\section{Acknowledgments}
This work was produced as part of the activities of FAPESP Research, Disseminations and Innovation Center for Neuromathematics (Grant 2013/07699-0, S. Paulo Research Foundation). Morgan André is supported by a FAPESP scholarship (grant number 2017/02035-7), Cecilia Romaro (grant number 88882.378774/2019-01) and Fernando Araujo Najman (grant number 88882.377124/2019-01) are the recipient of PhD scholarships from the Brazilian Coordenação de Aperfeiçoamento de Pessoal de Nível Superior (CAPES).

\bibliographystyle{acm}
\bibliography{Bibliografia}
\end{document}